# Gradient Based Hybridization of PSO

ARUN K PUJARI

Department of Computer Science & Engineering,

Ecole Centrale School of Engineering, Mahindra University, Hyderabad, India

arun.pujari@mahindrauniversity.edu.in

SOWMINI DEVI VEERAMACHANENI

Department of Computer Science & Engineering,

Ecole Centrale School of Engineering, Mahindra University, Hyderabad, India

sowminidevi.v@mahindrauniversity.edu.in

Particle Swarm Optimization (PSO) has emerged as a powerful metaheuristic global optimization approach over the past three decades. Its appeal lies in its ability to tackle complex multidimensional problems that defy conventional algorithms. However, PSO faces challenges, such as premature stagnation in single-objective scenarios and the need to strike a balance between exploration and exploitation. Hybridizing PSO by integrating its cooperative nature with established optimization techniques from diverse paradigms offers a promising solution. In this paper, we investigate various strategies for synergizing gradient-based optimizers with PSO. We introduce different hybridization principles and explore several approaches, including sequential decoupled hybridization, coupled hybridization, and adaptive hybridization. These strategies aim to enhance the efficiency and effectiveness of PSO, ultimately improving its ability to navigate intricate optimization landscapes. By combining the strengths of gradient-based methods with the inherent social dynamics of PSO, we seek to address the critical objectives of intelligent exploration and exploitation in complex optimization tasks. Our study delves into the comparative merits of these hybridization techniques and offers insights into their application across different problem domains.

## 1 INTRODUCTION

Numerous methods have been developed in mathematical programming (MP) for optimizing unconstrained nonlinear function, with many of them centered around the selection of a search direction that gradually guides the optimization process toward the optimal point. Determining these search directions often necessitates the computation of derivatives, either first or second order, related to the objective function and design constraints. A drawback of this branch of optimization lies in its susceptibility to converging to local optima.

Alternatively, there exist metaheuristic approaches that offer a distinct advantage: the ability to break free from local optima. Within the realm of population-based metaheuristics, there is a subset inspired by the collective intelligence observed in swarms. Notable examples of such swarm intelligence-based methods include Particle Swarm Optimization (PSO) [1], Ant Colony Optimization [2], Firefly Algorithm, Cuckoo Search, Artificial Bee Colony (ABC), Bat Algorithm, Differential Evolution (DE).

PSO stands out as highly suitable for optimizing multi-modal unconstrained nonlinear functions. Recognizing the advantages and disadvantages associated with gradient-based methods and swarm intelligence-based methods, researchers endeavor to synthesize these principles, exploring hybrid techniques that promise more effective solutions to nonlinear optimization problems.

The motivation behind the hybridization of Particle Swarm Optimization (PSO) and gradient-based optimization methods typically arises from the desire to leverage the strengths of both approaches while mitigating their respective weaknesses. By combining the two, we can benefit from global search capabilities to escape local optima and the use gradient-based techniques to refine the solution once a promising region is identified. Hybridization can speed up convergence, reducing the time needed to find a high-quality solution. Some hybrid approaches use adaptive techniques to switch between PSO and gradient-based methods during the optimization process. This adaptability can be beneficial in cases where the problem landscape changes over time. Hybrid approaches can incorporate domain-specific knowledge to guide the optimization process, which can be especially valuable in scientific or engineering applications. There are other benefits such as constraints handling, multi-objective optimization, robustness, scalability or optimal use of computational resources. However, it is important to note that the success of hybridizing PSO with gradient-based methods depends on the specific problem at hand, and the design strategy of the hybrid algorithm. In simple terms, the justification to combine these two paradigms is due to the following reasons.

1. **Search Diversification:** PSO focuses on global exploration and search diversification, attempting to discover promising regions of the search space. It leverages the swarm intelligence and social interactions among particles to explore different areas of the search space.
2. **Local Refinement:** Gradient descent, on the other hand, performs local exploitation and fine-tuning. It converges rapidly to local optima using gradient information of first or second order.

Notably, despite the considerable potential in the fusion of PSO and gradient-based search methods, there exists a gap in the literature—a comprehensive and systematic exploration of this subject is notably absent. It would be unjust to claim a complete absence of attempts in this regard. While some hybridization endeavors have been made, they often focus on specific applications without providing a clear rationale for the chosen application or the necessity for hybridization. Moreover, various proposals have emerged, aiming to merge these two paradigms in the context of general optimization problems. However, each study tends to introduce its unique hybridization scheme. Some efforts seek to enhance one of the methods by integrating it with the other. For example, in cases where gradient-based search traditionally prevails, attempts are made to hybridize it with the aim of elevating its performance. Nevertheless, the question of whether this hybrid method surpasses the effectiveness of pure PSO remains unanswered. Similarly, for applications where PSO is conventionally employed, improvements are sought by incorporating gradient information. Multiple hybridization schemes have been presented, yet it remains unclear which scheme holds greater justification in a given context. It is these observations that motivate the present study, aiming to address the existing gap and contribute to a more systematic exploration of the integration of PSO and gradient-based search methods in the realm of optimization. To the best of our knowledge, there has not been any earlier attempt in this direction and we believe that the present work will facilitate to in establishing a new direction of scientific investigation and enquiries.

The paper is structured as follows. In Section 2, we provide an overview of PSO and gradient-based search to establish a common notation for both frameworks and ensure completeness. Section 3 formalizes various hybridization strategies, specifically categorizing them as coupled, decoupled and concurrent hybridization. The sequential (or decoupled) hybridization has minimal impact on the base algorithms, primarily involving their appropriate sequencing to create a hybrid version, while the coupled hybridization incorporates proposals that significantly affect the base algorithms. We focus more on the coupled hybridization in Section 4, with a detailed examination, and also identify the hybridization processes for specific applications as well as for general optimization. In Section 5, the sequential hybridization is discussed. In Section 6, we analyze the experimental results for different hybridization schemes. Section 7 discusses future research suggestions and provides the conclusions.

## 2 BASICS OF PSO AND GRADIENT-BASED OPTIMIZATION

In this section the background information relating to PSO and the gradient-based search are developed.

### 2.1 Particle Swarm Optimization

PSO is a swarm intelligence-based metaheuristic technique that searches for the optimal solution by simulating the movement of birds in a flock. The population of the birds is called swarm, and the members of the population are particles. Each particle represents a possible solution to the optimizing problem. During each iteration, each particle flies independently in its own direction, which is guided by its own previous best position as well as the global best position of all the particles. Assume that the dimension of the search space is D, and the swarm is $S = (X_1, X_2, \ldots, X_{Np})$; each particle represents a position in the D dimension; the position of the i[th] particle in the search space can be denoted as $X_i = (x_{i1}, x_{i2}, \ldots, x_{iD}), i = 1, 2, \ldots, Np$, where *Np* is the number of all particles. The own previous best position of the i[th] particle is called *pbest* which is expressed as $P_i = (p_{i1}, p_{i2}, \ldots, p_{iD})$. The best position of all the particles is called *gbest*, which is denoted as $P_g = (p_{g1}, p_{g2}, \ldots, p_{gD})$. We introduce suffixes to $X_i$, $P_i$ and $P_g$ to refer to these quantities at time *t*, e.g., $X_i^t$, $P_i^t$ and $P_g^t$. In the same manner, the velocity of the i[th] particle at time *t* is expressed as $V_i^t = (v_{i1}{}^t, v_{i2}{}^t, \ldots, v_{iD}{}^t)$.

In the standard PSO [1] several design vectors (particles) are randomly generated in the design space of the unconstrained function *f(X)*. Then the positions of particles are randomly updated based on the following equations:

$$V_i^{t+1} = \omega V_i^t + c_1 \times rand() \times (P_i^t - X_i^t) + c_2 \times rand() \times (P_g^t - X_i^t) \qquad (1)$$

$$X_i^{t+1} = X_i^t + V_i^{t+1} \qquad (2)$$

where $c_1$ and $c_2$ are the acceleration constants with positive values; and rand() is a random number ranging from 0 to 1.

The inertia weight $\omega$ is a crucial parameter in Particle Swarm Optimization (PSO). Inertial weight plays a pivotal role in balancing the trade-off between exploration and exploitation in PSO. A high weight value promotes exploration, and a low weight value promotes exploitation. A high inertial weight allows particles to move quickly and cover a broader area of the search space, facilitating the discovery of promising regions or global optima particularly for multimodal functions. A low inertial weight is beneficial when the search space is well-sampled, and particles need to refine their positions around known optima.

*Constriction coefficient*, also known as *constriction factor*, is another key component in many PSO variants and plays a critical role in the algorithm's convergence behavior. The velocity update equation with the constriction coefficient is as follows:

$$V_i^{t+1} = \Phi \times [V_i^t + c_1 \times rand() \times (P_i^t - X_i^t) + c_2 \times rand() \times (P_g^t - X_i^t)] \qquad (3)$$

where, Φ (constriction coefficient) is a parameter that controls the effect of the inertia and cognitive and social components on the velocity update. The constriction coefficient ensures that the particles' velocities are bounded and that they neither move too fast nor become too stagnant. It also helps in controlling the trade-off between exploration (global search) and exploitation (local search) during the optimization process. Various formulations of PSO exhibit differences in their choices between weighted inertia and constriction coefficients, as well as in the methods they employ to calculate these parameters within the algorithm.

## 2.2 Gradient-Based Optimization for Unconstrained Nonlinear Function

Gradient descent is a way to minimize an objective function $f(X)$, $X \in R^D$ by searching along the negative direction of the gradient, $-\nabla_X(f(X))$ which is usually called steepest descent direction. The plain "gradient descent method" [3] to find the minimum of $f(X)$ starts from an initial point $X_0$ in $R^D$ then iteratively takes a step along the steepest descent direction (optionally scaled by a stepsize), until convergence. The stepsize or the learning rate η determines the size of the steps which is computed in each iteration adaptively either by line search, by backtracking search or by some rule. The update rule for the gradient descent is as follows.

$$X_i^{t+1} = X_i^t + \eta^t \times \left(-\nabla(f_{X_i^t})\right) \tag{4}$$

There are methods like Newton's Method [3] which make use of second derivatives unlike gradient search which uses the first order derivatives. For a quadratic function (with positive definite $H$) Newton's method directly jumps to the minimum. The update rule for Newton's Rule is given by

$$X_i^{t+1} = X_i^t + \eta^t \times \left(-H^{-1}(f_{X_i^t})\nabla(f_{X_i^t})\right), \tag{5}$$

where H is the Hessian of $f$ at $X_i^t$.

While second-order derivatives offer efficiency by providing a one-shot solution for quadratic functions, the computational cost and the potential challenges of obtaining the Hessian matrix, especially in closed form or when it is non-invertible, can be significant. As a result, approaches such as the Quasi-Newton method [4] aim to emulate Newton's method by creating approximations of the Hessian using easily computable functions. The primary goal of Quasi-Newton methods is to produce a sequence of approximate Hessian matrices denoted as $H_0, H_1, ...$ such that the $t^{th}$ matrix only depends on the $(t-1)^{th}$ matrix and satisfies the desired constraint of approximation. Quasi-Newton methods constitute a class of optimization techniques, with each method specifying its unique approach to approximating the Hessian matrix. Among these methods, BFGS (Broyden-Fletcher-Goldfarb-Shanno) stands out as the most widely recognized and utilized. The update rule is given by

$$X_i^{t+1} = X_i^t + \eta^t \times \left(-B(f_{X_i^t})\nabla(f_{X_i^t})\right), \tag{6}$$

where B is a matrix that approximates the inverse of the Hessian.

The Conjugate Gradient (CG) [4] method combines elements of both gradient descent and the method of conjugate directions. It aims to find the optimal solution by iteratively moving in conjugate directions, which are directions in the search space that are orthogonal (uncorrelated) to each other with respect to a specific matrix. The primary advantage of the Conjugate Gradient method is that it can converge to the optimum in a relatively small number of iterations, making it efficient for solving large problems. The update rule is given by

$$X_i^{t+1} = X_i^t + \eta^t \times (p_i^t) \tag{7}$$

where $p_i^t = -\nabla(f_{X_i^t}) + \beta_i^{t-1} p_i^{t-1}$,  $\beta_i^{t-1} = \frac{\left|\nabla(f_{X_i^t})\right|^2}{\left|\nabla(f_{X_i^{t-1}})\right|^2}$ and step length $\eta^t = \frac{\nabla(f_{X_i^t})^T \nabla(f_{X_i^t})}{(p_i^t)^T H p_i^t}$

Gradient Descent and Particle Swarm Optimization (PSO) are both optimization techniques used for unconstrained optimization problems and there are some pros and cons of each method. Gradient Descent is highly efficient for smooth and well-behaved functions, especially in high-dimensional spaces. It follows a clear and deterministic path towards the optimal solution, which can be advantageous in certain scenarios. In convex optimization problems, Gradient Descent is guaranteed to converge to the global minimum. However, it can be sensitive to the initial starting point, and might get stuck in local minima in non-convex problems. Gradient Descent can perform poorly when dealing with noisy or non-smooth functions. It relies on the availability of derivatives (gradients), which can be a limitation in some situations. PSO is good at exploring the entire search space and can often find global optima in complex, non-convex problems. It doesn't rely on derivatives, making it applicable to a wider range of problems, including non-smooth and discontinuous functions. PSO operates with a population of solutions, which can help in escaping local optima by sharing information among particles. PSO's stochastic nature can lead to variability in results and make it less predictable compared to Gradient Descent. It may prematurely converge in some cases, and there is no guarantee of global convergence. PSO requires tuning of parameters like the inertia weight, acceleration coefficients and constriction coefficient, which can be challenging. The choice between these methods depends on the specific characteristics of the optimization problem at hand.

## 3 HYBRIDIZATION PRINCIPLES

Integrating gradient descent with Particle Swarm Optimization (PSO) can result in the creation of hybrid optimization techniques that harness the global search capabilities of PSO and the local refinement provided by gradient descent. Various endeavors have been made to amalgamate these two paradigms for addressing both general nonlinear optimization problems and specific applications. This article seeks to establish a formal framework for hybridization, presenting a range of methods and strategies for combining these two optimization approaches:

**Coupled Hybridization or Gradient Information Incorporation**: Integrating gradient information into the velocity or position updates of particles within the PSO algorithm involves merging the PSO update equation with gradient-based updates. The standard PSO update equation comprises three essential terms (Eq 1): inertia, the term associated with the local best, and the term associated with the global best. These three terms are collectively weighted using random parameters, which are often referred to as constriction coefficient or are separately weighted with inertial weights. Hybridization approaches diverge in how the gradient term is incorporated. Some approaches introduce the gradient term as an additional component, enhancing the traditional three-term equation. In contrast, others opt to replace one of the existing terms in the equation with gradient-based information, modifying the balance of the algorithm.

**Sequential Hybridization**: In this hybridization approach, Particle Swarm Optimization (PSO) and gradient-based search are sequentially employed, with PSO often being the initial step, followed by gradient search. The process commences with global optimization using PSO. When PSO reaches a specific level of convergence or approaches a promising solution, the algorithm transitions to gradient descent for local refinement. During this phase, gradient information steers particles or individuals toward the nearest local optima. Adaptive mechanisms are integrated to facilitate the transition between PSO and gradient descent, contingent upon specific criteria like the objective function's value or the diversity within the swarm. Some researchers have also proposed sequential process with first a gradient-based search followed by a PSO. In such a situation, generating swarm in the middle of the algorithm is an issue.

**Concurrent Hybridization**: In this mode of hybridization, PSO and gradient descent are activated simultaneously, with each algorithm operating independently. The best solution found by either algorithm can be shared with the other to potentially influence its search. In concurrent hybridization, both PSO and gradient descent algorithms are executed in parallel, typically on separate processors or threads.

While these three approaches offer diverse possibilities, most research primarily concentrates on the first two: *sequential (decoupled)* and *coupled* hybridization. Variations arise in other aspects within these major schemes, such as the allocation of different weights to PSO and gradient descent. Some methodologies adopt variable weighting, with weight assignments evolving throughout the optimization process. This often begins with a higher weight for PSO to emphasize global exploration and gradually shifts towards a higher weight for gradient descent to emphasize local exploitation. The degree of collaboration and information sharing between PSO and gradient descent can also fluctuate based on problem characteristics and algorithm performance. We review the existing methods in the following sections.

## 4 COUPLED HYBRIDIZATION

In this section, the major methods of incorporating gradient information within the update equation of PSO are discussed.

The coupled hybridization method introduced by Maeda et al. [5] distinguishes itself from other approaches due to its unique characteristic: replacing the conventional weighted inertia with a gradient approximation. In most hybrid methods (as we see in this section), gradient information is typically introduced as an additional component or as a substitute for the local-best direction within the PSO algorithm. However, in this approach, the gradient approximation takes precedence over the traditional weighted inertia, setting it apart from conventional hybridization strategies. Additionally, the authors explore several schemes of implementation, Scheme 1: Simultaneously updating all particles; Scheme 2: Updating only the best particles and Scheme 3: Updating half of the population of particles. The paper conducts a comparative analysis against traditional PSO using various test functions of De Jong test suite [6] and neural network learning problems. Through this analysis, it provides insights into the strengths and weaknesses of the proposed simultaneous perturbation particle swarm optimization across different schemes. It is worth noting that, in the case of test functions, the hybrid PSO outperforms traditional PSO.

The approach presented in [7] combines elements of both coupled and decoupled methods. A notable feature of this method is its adaptability to the current diversity level within the swarm. When diversity within the swarm is high, it harnesses the power of traditional PSO with weighted inertia. In contrast, when diversity decreases, it transitions into a coupled hybrid PSO with gradient-based components. This approach, referred to as DGPSOGS, integrates various factors into its velocity term, including weighted inertia, normalized negative gradient direction, and a diversity-damped direction based on the global best. The update equation, given below, illustrates this aspect of the formulation.

$$V_i^{t+1} = \omega V_i^t + c_1 \times rand() \times \left(\frac{-\nabla f_{X_i^t}}{\left\|\nabla f_{X_i^t}\right\|}\right) + c_2 \times rand() \times \left(\frac{(P_g^t - X_i^t)}{diversity(t) + \varepsilon}\right) \qquad (8)$$

where, $diversity(t) = \frac{1}{N_p \times |L|} \sum_{i=1}^{N_p} \sqrt[2]{\sum_{j=1}^{D}(x_{ij}^t - \bar{x}_j^t)^2}$ [4] and $\varepsilon$ is a predetermined small positive number to avoid the denominator becoming zero.

The research involved conducting experiments on five distinct test cases of De Jong test suite [6], each varying in dimension. These results were then compared against the performance of five different PSO variants. The findings highlight that the proposed algorithm not only converges to lower objective function values but also achieves faster convergence while effectively managing diversity within the swarm.

In the context of Recommender Systems, Sowmini Devi et al. [8] introduced a coupled hybrid PSO approach PSO-MMMF for Maximum Margin Matrix Factorization, a technique utilized in recommender systems. The velocity update equation is given as follows:

$$V_i^{t+1} = V_i^t - c \times \left(\delta \times (-\nabla f_{X_i^t}) + (1-\delta) \times (P_g^t - X_i^t)\right) \quad (9)$$

where, $\delta$ (0 ≤ $\delta$ ≤ 1) is the weight for the gradient search. In this setup, inertia is not weighted, and the term related to the local best is disregarded. Additionally, a constant represented by *c* serves as the constriction coefficient in this context. The conventional Particle Swarm Optimization (PSO) method is modified by replacing the component associated with the local best with the negative gradient term in the velocity updates.

Using the above formulation, Salman et al. [9] investigated various PSO topologies aimed at reintegrating the global best term. The search direction is a combination of the gradient direction and an adjustment favoring swarm exploration, influenced by the global best particle. The study considered different topologies, including fully connected, star, ring, and Von-Neumann topologies, to assess their impact on the algorithm's ability to balance exploration and exploitation.

Subsequently Laishram et al. [10] study PSO-MMMF further to observe that diversity plays a role in the efficacy of the method. Two more variants of PSO-MMMF [8] are proposed. In the first variant, the swarm diversity is intermittently increased by rerandomizing some particles in the swarm by GA-inspired methods. In the second variant, termed as HPSO-MMMF the neighborhood information is controlled by arranging the particles structurally rather than assigning particles randomly. However, the fundamental concept of hybridization proposed in [8] remains consistent throughout these adaptations.

In the realm of medical diagnostics, Yadav et al [11] examined the potential benefits of hybridizing the training procedure for Artificial Neural Networks (ANNs) with swarm intelligence. The authors introduce PSO-BP, a hybrid PSO approach that essentially employs a similar strategy to integrate a gradient term into the velocity updates as an additional fourth term.

$$V_i^{t+1} = \omega V_i^t + c_1 \times rand() \times (P_i^t - X_i^t) + c_2 \times rand() \times (P_g^t - X_i^t) - \eta \nabla f_{X_i^t} \quad (10)$$

Salajehgheh et al. [12] propose a coupled hybridization approach focused on enhancing the effectiveness of Particle Swarm Optimization (PSO) in design problems. In their method, they extend the capabilities of the PSO algorithm by incorporating both first-order (PSOG1) and second-order gradient directions (PSOG2) into the velocity updates of the particles. In PSOG, the gradient direction is first normalized and scaled with the size of the PSO vector. Then the vectors generated by PSO and gradients are added through the scalar parameters $c_3$ and $c_4$ by a random nature, $r_4$. The process is formalized as follows.

$$V_i^{t+1} = c_3 \times V_{i-pso}^t + c_4 \times r_4 \times \left(\frac{-B_i^t \nabla f_{X_i^t}}{\left|B_i^t \nabla f_{X_i^t}\right|}\right) \times \left|V_{i-pso}^t\right| \quad (11)$$

where $B$ is the approximate Hessian matrix and $V_{i-pso}^t$ is the velocity obtained Eq 1.

In PSOG1, each PSO particle, generated randomly, is equipped with two search directions: one generated by PSO and another along the negative gradient direction of the function under optimization. These two directions are harmonized using scaling factors to create a more efficient and dependable direction for the particles. The idea of variable metric method is employed to enhance POSG1 to PSOG2. Their approach takes advantage of both first and approximate second-order gradients, and they employ a penalty function method to convert constrained optimization problems into unconstrained ones. Design problems drawn from the existing literature serve as the basis for comparing their hybrid approach with the original PSO method. To demonstrate the improved performance of their novel approach, the authors conducted experiments using examples from benchmark functions and structural design problems. Interestingly, this study indicates that second-order derivative is no better than the first-order when combined with swarm intelligence.

Liu et al. [13] conducted a study focusing on optimization algorithms for neural network parameter tuning. They examined combining the stochastic gradient descent with momentum (SGD) with particle swarm optimization (PSO) and proposed PSO-SGD algorithm, a hybrid approach combining gradient descent with momentum and particle swarm optimization. The velocity update equation for PSO-SGD, a straight forward incorporation gradient information as an additional term (Eq 10). The algorithm's performance was assessed using ResNet18 for image classification on two datasets: the Blood Cell Images Data Set [14] and COVID-19 Radiography Data Set [15]. It is shown that PSO-SGD outperforms traditional PSO as well as traditional SGD (Stochastic Gradient Descent).

Davi et al. [16] introduced an innovative approach called PSO-PINN. This method addresses the challenges of training Physics-informed neural networks (PINNs) using standard gradient descent, which often faces non-convergence issues, especially with partial differential equations (PDEs) featuring irregular solutions. To tackle these issues, the paper proposes a novel approach that leverages Particle Swarm Optimization (PSO) for training PINNs. The PSO-PINN algorithm effectively mitigates the common problems associated with PINNs trained using traditional gradient descent. The paper also presents PSO-BP-CD, a hybrid variant combining PSO and gradient descent, with the gradient-based search gaining more importance as the training progresses and approaches a promising local optimum. The paper used the same formulation (Eq 10) as in Liu et al [13] and Yadav et al [11]. In addition, iteratively, the behavioral coefficients $c_1$ and $c_2$ decrease linearly as $c_k^{t+1} = c_k^t - \frac{2c_k^t}{n}$, where $k = 1, 2$ and $n$ is the number of iterations.

Zou et al [17] delved into the integration of Particle Swarm Optimization (PSO) with gradient-based techniques in a different application domain-the well-logging-constrained impedance inversion process for anticipating coal seam thickness and bifurcation. In essence, their paper combines the conjugate gradient method with Particle Swarm Optimization (PSO) to achieve highly accurate and reliable predictions of coal seam thickness and bifurcation. The results obtained with the CG-PSO algorithm outperformed the conventional statistical wavelet pickup method, establishing it as the preferred choice for wavelet inversion operations.

## 5 SEQUENTIAL HYBRIDIZATION

In this section, the major methods which consider the PSO and Gradient-based search separately but used in sequence are discussed.

Noel [18, 19] proposed a sequential scheme of hybridization, namely gradient-based PSO (GPSO) which uses a mild deviation from the traditional PSO methods by eliminating inertial weights and constriction coefficients. In the first phase, GPSO employs PSO to locate favorable local minimum, and subsequently, it employs the quasi-Newton-Raphson (QNR) algorithm to search for the global minimum with the best solution found so far as the starting point. The best solution found by the QNR search then becomes the best solution for the next PSO iteration, and the algorithm repeats until termination at a preset number of repetitions. The hybrid algorithm was benchmarked on the De Jong test suite [6] of benchmark optimization problems and compared to a hybrid algorithm that used a standard PSO. The GPSO variant achieved a more accurate solution and converged faster.

Chen et al. [20] introduced a sequential hybrid PSO algorithm that leverages the advantages of traditional PSO and gradient-based local search techniques. It initiates with global optimization using conventional PSO, aiming to find an approximate local minimum. Subsequently, it introduces a conjugate gradient-based search process that utilizes the best solution identified during the initial PSO phase. The results of simulation experiments confirm the algorithm's ability to achieve faster convergence when compared to genetic algorithms (GA). It also demonstrates superior performance in the parameter identification of radial basis function (RBF) neural networks.

Sequential hybridization approaches typically commence with Particle Swarm Optimization (PSO) as the initial step, subsequently followed by a gradient-based search. Nonetheless, in [21], the authors introduced GRPSO, a distinctive approach in which gradient-based methods took the lead role initially to direct the search towards locating a local minimum. Subsequently, the PSO algorithm was used to guide particles in departing from the previously converged local solution in favor of a superior point, from which gradient methods could recommence. This process was reiterated until the termination condition was satisfied. To initialize the population for the PSO algorithm, a Gaussian distribution was employed. Furthermore, to leverage the acquired information for preventing premature convergence of the population, a repulsion technique was employed. In this study, the authors applied an update equation incorporating inertial weight. The experimental analysis involved comparisons with a gradient descent algorithm featuring dynamic tunnelling and the GPSO algorithm proposed by Noel [18, 19]. The results revealed that GRPSO achieved a remarkable success rate of 100%, whereas GPSO exhibited a significantly lower success rate, approximately around 20%.

Plevris et al [22] examined a hybrid approach involving the fusion of Particle Swarm Optimization (PSO) and a gradient-based quasi-Newton Sequential Quadratic Programming (SQP) method to address structural optimization problems. The central goal of this approach is to effectively navigate the design space and accelerate the process of converging towards the global optimum. This paper adopts a two-phase decoupled method, commencing with the conventional PSO approach and then transitioning to a GD-PSO hybrid method. The effectiveness of this hybrid approach is demonstrated through its application to a range of benchmark structural optimization problems, validating its efficiency in optimizing complex structural designs.

In [23], the sequential hybridization is addressed in the context of antenna design with an objective to optimize the design and to fulfill the sensitivity criterion simultaneously. In this proposal the particles in the neighboring area of the global optimum detect the gradient of the global optimum and determine the sensitivity of the solution. When PSO converges, the gradient search takes over. During the gradient descent, the sensitivity of the optimum and the movement of the particles are determined by gradient of the optimum particle. Gradient of the optimum solution is calculated by the arranged crosshair. The method proposed here appears to be interesting and novel, but no experimental analysis is reported.

The study conducted by Wessel and Van der Haar [24] explores the possibility of integrating Particle Swarm Optimization (PSO) with gradient-based methods for optimizing convolutional neural networks (CNNs). Their research involves a comparison of the performance between well-known gradient-based methods, such as stochastic gradient descent with and without momentum, and Adaptive Moment Estimation (Adam), both with and without the incorporation of PSO. The findings suggest that, except for networks optimized with Adam, conventional gradient-based methods generally outperform those combined with PSO in terms of various performance metrics. It is noteworthy that using PSO to refine solutions obtained through gradient descent results in more significant reductions in loss compared to using PSO as the initial optimization method. This study sheds light on the order in which these methods are employed in sequential hybridization and offers insights into their comparative effectiveness.

## 6 EMPIRICAL INSIGHTS AND PERFORMANCE EVALUATION

On this survey of the hybridization of PSO with gradient search, we recognize that theory and practical validation go hand in hand. This section serves as a repository of experimental results. These results stand as evidence of the efficacy and real-world applicability of the hybrid algorithms outlined in foregoing discussion. They provide insights into the performance improvements and capabilities of these hybrid approaches. Through this compilation of empirical findings, we aim to shed light on the impact and advancements achieved through the integration of PSO with gradient search, guiding the way for researchers and practitioners seeking optimized solutions across a spectrum of domains.

**TABLE I:** Compilation of Empirical Analyses

The first column includes the method/author names with references and specifies the application area if applicable. The second and third columns provide abbreviated descriptions of the datasets utilized and the sizes of the data employed for experiments, respectively. In column 4, the improvements resulting from hybridization are documented. Column 5 lists the method with which the hybrid method is compared. Column 5 summarizes the comparison of each hybrid method with one of the base algorithms, either PSO or GD. Column 6 identifies whether the hybridization is sequential (Sq) or coupled (C).

| Method & application | Data used 2 | Problem size 3 | Observation 4 | 5 | 6 |
|---|---|---|---|---|---|
| Maeda et al [5] ANN | DJTS (5 function) | 10 | Faster convergence | PSO | C |
| DGPSOGS [7] | DJTS(5 function) | 10, 20, 30 | -Faster convergence<br>-Better diversity | PSO | C |
| PSO-MMMF [8] Rec. Systems | MovieLens | 10K, 50K, 100K | -Lower function value<br>-Faster convergence<br>-Better accuracy | GD | C |
| HPSO-MMMF [9] Rec. Systems | MovieLens | 10K, 50K, 100K | -Better accuracy | GD | C |
| PSO-Global, PSO-Ring [10] Rec. Systems | MovieLens | 100K | -Faster convergence<br>-Lower function value | GD | C |
| PSO-BP [11] Med.Diagnosis | UCI | - | -Faster convergence<br>-Better accuracy | GD | C |

| Method & application | Data used | Problem size | Observation | 5 | 6 |
|---|---|---|---|---|---|
| PSOG [12] Structural Optimization | CEC 2017, CEC 2015 | 10, 30 | -Better accuracy -Faster computation | PSO | C |
| PSO+SGD [13] Med Image | BCIDS, COVED19RDS | 12,500 images, 21,165 images | -Faster convergence -Better accuracy | -GD -PSO | C |
| PSO-BP-CD [16] | Poisson Equation, PDE - Advection, Heat, Burgers, Forced Heat, The Allen-Cahn | --- | -Less $L_2$ error | PSO | C |
| HGPSO [18] | DJTS (6 functions) | --- | -Faster Convergence | PSO | Sq |
| GPSO [19] | DJTS- (5 functions) | 10, 20, 30 | -Faster convergence | PSO | Sq |
| PSO-SQP [22] Structural Optimization | 10 bar plane truss, 25 bar space truss, 72 bar space truss | --- | -Faster convergence -Lower function value | PSO | Sq |
| Wessels et al [24] CNN | MNIST | 70,000 images | -Better accuracy | SGD | Sq |

Table I presents a comprehensive summary of all experimental findings. Hybrid algorithms are classified based on their nature, either coupled or sequential, the datasets they utilize, the type of problem they tackle, whether general optimization or a specific application, and the outcomes they achieve, including enhanced convergence speed, lower function values, reduced computation time, or improved accuracy. To enhance readability, data sources are denoted in abbreviated form, and the specific details are provided below.

*DJTS* **[6**] – The De Jong Test Suite comprises a collection of mathematical functions, each with well-known optimal solutions, used for evaluating and comparing optimization algorithms.

*MovieLens* **[25]** data is a standard benchmark dataset in the field of Recommendation Systems It consists of user-generated movie ratings and movie metadata.

The *UCI* [26] Machine Learning Repository consists of a diverse collection of datasets designed for various machine learning and data science tasks.

**CEC 2015** [27] and **CEC 2017** are specialized competitions held in the framework of the IEEE Congress on Evolutionary Computation (CEC). These competitions are specifically organized to assess the performance of optimization algorithms, with a particular focus on those that utilize evolutionary computation methods.

*MNIST* **[28]** is a well-known dataset in machine learning and computer vision, primarily utilized for image classification. It consists of a collection of grayscale images of handwritten digits (0 to 9).

*BCIDS* **[14]** - Blood Cell Images dataset contains 12,500 enhanced blood cell images.

*COVID-19RDS* **[15]** – COVID-19 Radiography data set is a chest X-ray image data set, which currently has 21,165 images.

*10 bar plane truss, 25 bar space truss, 72 bar space truss* - These trusses are essential in structural engineering, providing the framework for various constructions.

*PDEs* **[29]** - Partial differential equations (PDEs) play a vital role in understanding and simulating physical systems and materials under different conditions.

It is apparent that each hybridization proposal claims superiority over one of the conventional algorithms, either PSO or GD. This observation remains consistent even when addressing specific application domains. However, it is important to note that each proposal employs different datasets obtained from diverse sources. Consequently, it becomes imperative to facilitate a fair comparison on a shared platform using common datasets and to assess their performance against both traditional base algorithms.

When examining the research findings related to PSO, a considerable amount of in-depth analysis has been dedicated to factors like inertial weight, constriction coefficient, particle diversity, and random particle generation. Likewise, the extensive body of literature concerning the advancement of gradient-based search methods has established theories for employing techniques such as the Conjugate Gradient Method, quasi-Newton method, and even Stochastic Gradient Descent.

However, in contrast, the field of hybridization techniques has not sufficiently explored the reasons or justifications for incorporating these specific methods into the hybridization process. There is a pressing necessity to delve into these aspects and conduct a thorough investigation.

## 7 CONCLUSIONS

In this study, we report a systematic exploration of the integration of Particle Swarm Optimization (PSO) and gradient-based optimization methods, with the primary goal of leveraging the strengths of both paradigms while mitigating their respective weaknesses. We strongly believe that the global exploration capabilities of PSO and the local refinement potential of gradient-based methods as two complementary facets that, when combined, could lead to more effective solutions. Through the categorization of hybridization strategies and a detailed examination of coupled hybridization in particular, this study sheds light on the various ways in which PSO and gradient-based methods can be combined for different types of optimization problems. The survey of the literature revealed a gap in the field – a comprehensive and systematic study that systematically categorizes, evaluates, and compares different hybridization schemes. While some hybridization attempts exist, they often focus on specific applications without offering a clear rationale for hybridization, leaving open the question of whether hybrid methods surpass the effectiveness of pure PSO or gradient-based methods. This study prompts several open research problems to fill that void by providing a comprehensive and systematic analysis of these hybridization schemes. This study paves the way for a more systematic and informed approach to the integration of PSO and gradient-based optimization methods. We hope that this research will encourage and facilitate additional investigations into this exciting area of study, offering new directions and insights for the optimization community.